\RequirePackage{fix-cm}
\documentclass[a4paper,11pt]{article}

\usepackage{authblk}
\usepackage{amsfonts,amsmath,amssymb,amsthm}
\usepackage{algorithm,algorithmic}
\usepackage{paralist}
\usepackage{multirow}
\usepackage{booktabs}
\usepackage{url}
\usepackage{hyperref}
\usepackage{graphicx}
\usepackage{microtype}
\usepackage{url}
\usepackage{hyperref}
\usepackage{subcaption}
\usepackage{tikz}
\usetikzlibrary{positioning}
\usetikzlibrary{decorations.pathreplacing}
\newcommand{\N}{\mathbb{N}}

\newcommand{\NP}{\mathbf{NP}}

\newtheorem{definition}{Definition}

\newtheorem{theorem}{Theorem}

\tikzset{
    mybrace/.style={decorate,decoration={brace,aspect=#1}}
}

\providecommand{\keywords}[1]{\textbf{\textit{Keywords }} #1}

\begin{document}

\title{On the Difficulty of Evolving Permutation Codes}

\author[1]{Luca Mariot}
\author[2]{Stjepan Picek}
\author[3]{Domagoj Jakobovic}
\author[3]{Marko Djurasevic}
\author[4]{Alberto Leporati}
	
\affil[1]{{\small Cyber Security Research Group, Delft University of Technology, Mekelweg 2, Delft, The Netherlands} 
	
	{\small \texttt{l.mariot,@tudelft.nl}}}

\affil[2]{{\small Digital Security Group, Radboud University, PO Box 9010, 6500 GL Nijmegen, The Netherlands} 
	
	{\small \texttt{stjepan@computer.org}}}

\affil[3]{{\small Faculty of Electrical Engineering and Computing, University of Zagreb, Unska 3, Zagreb, Croatia}
	
	{\small \texttt{\{domagoj.jakobovic,marko.durasevic\}@fer.hr}}}

\affil[4]{{\small Dipartimento di Informatica, Sistemistica e Comunicazione, Università degli Studi di Milano-Bicocca, Viale Sarca 336/14, Milano, 20126, Italy}
    	{\small \texttt{alberto.leporati@unimib.it}}}

\maketitle

\begin{abstract}
Combinatorial designs provide an interesting source of optimization problems. Among them, permutation codes are particularly interesting given their applications in powerline communications, flash memories, and block ciphers. This paper addresses the design of permutation codes by evolutionary algorithms (EA) by developing an iterative approach. Starting from a single random permutation, new permutations satisfying the minimum distance constraint are incrementally added to the code by using a permutation-based EA. We investigate our approach against four different fitness functions targeting the minimum distance requirement at different levels of detail and with two different policies concerning code expansion and pruning. We compare the results achieved by our EA approach to those of a simple random search, remarking that neither method scales well with the problem size. 
\end{abstract}


\section{Introduction}
\label{sec:intro}

Permutation codes (also called permutation arrays) are a particular kind of error-correcting codes where the codewords are permutations. In particular, a permutation code $PA(n,d)$ is a set of permutations of length $n$ such that any two permutations in it disagree in at least $d$ positions.

These combinatorial objects have several applications, for example, as error-correcting codes in \emph{powerline communications}~\cite{chu04}. The basic approach for powerline transmission is to encode the data by small voltage variations, with the requirement of keeping the power output as constant as possible. Further, transmission over powerlines is affected not only by white Gaussian noise but also by impulse and narrow-band noise due to electrical interference and magnetic fields. Using permutation codes in a modulation scheme, as suggested by Han Vinck~\cite{hanvinck00}, provides a good trade-off between power variation and correcting errors introduced by these kinds of noise.
A second domain where permutation codes have been extensively applied is \emph{flash memories}, particularly in the so-called \emph{rank-modulation scheme}~\cite{jiang08}. In traditional designs, the cells in a flash memory encode the information using different charge levels, allowing them to store a set of discrete values. On the other hand, rank-modulation encodes the data in the cells with a permutation that specifies the relative ranks of the charges instead of directly using their absolute values. Using a permutation code in this scheme improves the writing speed and the correction of errors introduced by charge leakage, which becomes progressively frequent in aging memories.

Finally, permutation codes have also been applied to a smaller extent in cryptography, specifically for the design of \emph{block ciphers}~\cite{delatorre00}. In the \emph{Substitution-Permutation Network} (SPN) paradigm for block ciphers, the plaintext is encrypted by iteratively applying several times a \emph{round function}. The round function, in turn, consists of a \emph{confusion layer}, which aims at making the relationship between the ciphertext and the symmetric key as complicated as possible, and a \emph{diffusion layer}, whose goal is to spread the statistical structure of the plaintext over the ciphertext. The resulting block is mixed with a \emph{round key} to get the corresponding ciphertext, which is then given as input to the next application of the round function. The diffusion layer is usually implemented through a \emph{Maximum Distance Separable (MDS) matrix} as it happens, for example, in AES~\cite{daemen20}. An alternative approach is to use a set of different permutations coming from a permutation code $PA(n,d)$. By dynamically choosing a different permutation from the code at each round, two different input blocks are guaranteed to result in output blocks at Hamming distance of at least $d$, thereby implementing a \emph{multi-permutation} as defined by Vaudenay~\cite{vaudenay94}.

Despite their simple definition, the construction of permutation codes is far from being a trivial problem. Indeed, finding the largest permutation code is a particular instance of the \emph{sphere-packing problem} studied in coding theory~\cite{conway88}, and of the {\sc Max-Clique} problem in graph theory, which is known to be $\NP-$complete~\cite{karp72}. In particular, one of the main open questions in this research field is to determine the largest permutation code for a given length $n$ and minimum distance $d$, i.e., the maximum number of permutations that can partake in a $PA(n,d)$. Such a number is usually denoted as $M(n,d)$, and its exact value is known only for a few specific cases. Generally, one resorts to coding-theoretic results to provide lower and upper bounds on $M(n,d)$. Apart from algebraic constructions, for which the reader may find a survey in~\cite{colbourn99}, a few heuristic algorithms have also been developed to construct large permutation codes~\cite{smith13,montemanni16}, mostly based on branch and bound and iterative clique search approaches. As far as we know, up to now, there have been no attempts in the literature to employ Evolutionary Algorithms (EA) to address this problem, although some authors used EA in the past to evolve other kinds of combinatorial designs, such as \emph{orthogonal Latin squares}~\cite{mariot17}, \emph{orthogonal arrays}~\cite{mariot18} and \emph{disjunct matrices}~\cite{knezevic18}.

This paper investigates the suitability of EA to optimize permutation codes. We do so by %considering two different optimization approaches: in the first, an EA directly searches for permutation codes in the space of subsets of permutations, while the second employs an EA to construct a large permutation code incrementally. We investigate in detail the latter approach, devising an iterative EA that starts from a random permutation and then evolves one permutation at a time until it finds one at a Hamming distance greater than or equal to the desired minimum
%\todo{Exactly that distance?}
from the previous permutations. The process is repeated until either a given fitness budget expires or an upper bound on $M(n,d)$ is reached (since this means that the code cannot be expanded further).
%\todo{This last sentence is not clear}
%\todo{let us shorten the paragraph and skip discussing the first approach as nothing is done with it?}
%\todo{a sentence what we achieved}
We evaluate our incremental EA approach under four fitness functions and two \emph{update policies}. The first update policy expands the code as soon as a suitable permutation is found by the EA. The second policy also removes some rows at random from the current code after a certain amount of fitness evaluations with no improvement has elapsed. The number of removed rows is decreased over time, similarly to the cooling schedule used in simulated annealing. For the sake of comparison, we also adopt a baseline random search (RS) method and perform experiments over 15 problem instances. The results show that both EA and RS cannot scale well on this optimization problem, with the largest codes found that lie far from the best-known lower bounds~\cite{smith12}. 
%We interpret the obtained results and discuss several ways to improve the findings reported in this paper. 

% The rest of this paper is structured as follows. Section~\ref{sec:bg} recalls all necessary background definitions and results about permutation codes. Section~\ref{sec:inc-constr} first discusses the "all-at-once" optimization approach for evolving permutation codes along with its limitations, and then describes in detail the incremental EA approach, together with the fitness functions defined for this problem. Section~\ref{sec:exp} describes an experimental plan to test the incremental EA on several problem instances.

\section{Preliminaries}
\label{sec:bg}

We denote by $[n]$ the set $\{1,\cdots,n\}$ of the first $n \in \N$ positive integers. Next, $S_n$ denotes the \emph{symmetric group} of order $n$, i.e., the set of all permutations over $[n]$. Given a permutation $\pi \in S_n$, we encode it as a vector $\pi = (p_1,\cdots, p_n)$ of length $n$, where each coordinate $\pi[i]$ specifies the value of the permutation when evaluated on $i \in [n]$. Further, given two permutations $\pi, \sigma \in S_n$, the \emph{Hamming distance} $d_H(\pi, \sigma)$ is the number of coordinates where $\pi$ and $\sigma$ differ.

%We begin by formally introducing the basic definition of \emph{permutation code}, which will be the main combinatorial object of interest considered in this paper.

\begin{definition}
	\label{def:pa}
	Let $n \in \N$ and $d \le n$. A \emph{permutation code} (also \emph{permutation array}, PA) of length $n$ and minimum distance $d$, denoted by $PA(n,d)$, is a subset $P$ of the symmetric group $S_n$ such that $d_H(\pi, \sigma) \ge d$ for every pair of distinct permutations $\pi, \sigma \in P$.
\end{definition}

%\todo{This sentence is redundant} In other words, a $PA(n,d)$ is a set $P$ in which any two permutations of $n$ elements differ in at least $d$ positions. 
Using the error-correcting codes terminology, the permutations in a $PA(n,d)$ are also called \emph{codewords}. If $P$ is composed of $m$ codewords, one can represent it through a $m\times n$ matrix where each row corresponds to one of the permutations in $P$. The ordering of the rows in such a matrix is irrelevant since it does not change the pairwise Hamming distances of the permutations. In the following, we will mostly use this matrix-based notation to represent $PA$, although the set-theoretic notation will also be useful to describe the operations performed by our evolutionary algorithms to search for such arrays.

% A concept closely related to PA is \emph{equidistant permutation codes}, which we define as:
% \begin{definition}
% \label{def:epa}
% An \emph{equidistant permutation code} (also \emph{equidistant permutation array}, EPA) of length $n$ and distance $d$, denoted by $EPA(n,d)$, is a subset $E$ of the symmetric group $S_n$ such that $d_H(\pi, \sigma) = d$ for each pair of distinct permutations $\pi, \sigma \in E$.
% \end{definition}
% Clearly, every $EPA(n,d)$ is also a $PA(n,d)$. Thus, equidistant PA is a much more constrained combinatorial object than simple PA since each pair of permutations must disagree in \emph{exactly} $d$ positions. Intuitively, given $n$ and $d$, one can conclude that there are significantly fewer EPA than PA. 

%As customary in the theory of error-correcting codes, o
One of the main problems is determining the largest number of codewords that can partake in a permutation code. Following the notation from~\cite{colbourn99}, given $n$ and $d$ we denote by $M(n,d)$ the maximum number of rows in a $PA(n,d)$. The values of $M(n,d)$ are generally unknown, but several theoretical bounds exist. Two well-known results in this direction, originating from coding-theoretical considerations, are the \emph{Gilbert-Varshamov lower bound} and the \emph{sphere-packing upper bound}, which we summarize below for permutation codes.

\begin{theorem}
	\label{thm:bounds}
	Let $n,d \in \N$ with $d\le n$. Then, the following inequalities hold for the maximum number of codewords in a permutation code:
	\begin{equation}
	\label{eq:bounds}
	\frac{n!}{\sum_{k=0}^{d-1} \binom{n}{k} D_k} \le M(n,d) \le \frac{n!}{\sum_{k=0}^{\left\lfloor\frac{d-1}{2}\right\rfloor} \binom{n}{k} D_k} \enspace ,
	\end{equation}
	where $D_k$ is the number of \emph{derangements} of $k$ elements (i.e., the number of permutations of length $k$ without fixed points), which for all $k \in \N$ equals:
	\begin{equation}
	\label{eq:der}
	D_k = k! \sum_{i=0}^k \frac{(-1)^i}{i!} \enspace .
	\end{equation}
\end{theorem}

The Gilbert-Varshamov and sphere-packing bounds are rather crude, but in practice, they are helpful to decide whether a specific instance of $n$ and $d$ is suitable to construct a permutation code large enough for a specific application. Of course, tighter bounds have been proved in the related literature of permutation codes, either by combinatorial arguments or by providing concrete constructions. The latter case usually occurs for specific values of the minimum distance. For instance, if $d=n$, then it is rather easy to constructively prove that $M(n,n) = n$, by considering any \emph{Latin square} of order $n$ as an example of permutation code reaching this bound. Indeed, a Latin square of order $n$ is a $n\times n$ array such that each number in $[n]$ appears exactly once in each row and each column. Thus, each row of the square is a permutation, and any two rows differ in all coordinates since there cannot be any repeated number in any column. Further, a simple construction for a Latin square of order $n$ is to take all \emph{cyclic shifts} of the identity permutation $(1,2,\cdots,n)$, which proves the existence of a $PA(n,n)$ for every $n \in \N$.

Latin squares also provide a construction for a better lower bound on $M(n,d)$ when $d = n-1$. In particular, two Latin squares are called \emph{orthogonal} if their superposition yields all ordered pairs in the Cartesian product $[n] \times [n]$, and a set of $k$ \emph{mutually orthogonal Latin squares} ($k$-MOLS) is a family of $k$ Latin squares of order $n$ that are pairwise orthogonal. Colbourn et al. ~\cite{colbourn04} showed how to construct a $PA(kn,n-1)$ by using a set of $k$-MOLS of order $n$, thereby proving that $kn \le M(n,n-1)$. %\todo{not sure what to do with this sentence? it starts as EPA, but it also gives results for PA?}
We emphasize that determining the maximum size of a MOLS family for a given $n$ is also a long-standing open problem in design theory, but several results are known for specific cases~\cite{stinson04}.

It is also easy to determine the maximum number of rows in a permutation array for low minimum distances. Indeed, two permutations cannot differ in only one position since this would imply that both vectors have a repeated value (thus, not making it a permutation). Therefore, the minimum distance is always at least 2, i.e., when two permutations differ by a single \emph{transposition}, or swap. This means that the largest $PA(n,2)$ corresponds to the symmetric group $S_n$ itself, hence $M(n,2) = n!$. 
%On the other hand, $M'(n,2) = 2$, since there cannot be more than two permutations that differ exactly by a single swap. 
Additionally, any two distinct permutations in the \emph{alternating group} $A_n$ (i.e., the set of \emph{even} permutations of length $n$) are always at a minimum distance of 3. Since the alternating group is exactly half of the size of the symmetric group, it follows that $M(n,3) = n!/2$. %todo{not sure what to do with this sentence?}
%With more involved arguments~\cite{bolton72}, one can finally show that $M'(n,3) = n-1$ for $n\ge 4$ and $M'(n,3) = \lfloor n/2 \rfloor$ for $n\ge 10$. 
We conclude this section by summarizing the above results as follows:
\begin{theorem}
	Let $n,d \in \N$ with $d\le n$. Then:
	\begin{compactitem}
		\item $M(n,1) = 1$ ; $M(n,2) = n!$; $M(n,3) = n!/2$; % $M'(n,3) = \begin{cases} n, & n<4, \\ n-1, & 4 \le n < 10, \\ \lfloor n/2 \rfloor, & n \ge 10, \end{cases}$
		\item $M(n,n) =  n$; $kn \le M(n,n-1)$, if there exists a set of $k$-MOLS of order $n$.
	\end{compactitem}
\end{theorem}
Tables reporting more refined lower and upper bounds for various values of $n$ and $d$ may be found in~\cite{smith12}.

% \section{Related Work}
% \label{sec:related}

% To the best of our knowledge, no prior works consider evolutionary algorithms to evolve permutation codes. 
% Still, there is a (limited) body of work where evolutionary algorithms are used to evolve different combinatorial designs.

%  \emph{orthogonal Latin squares}~\cite{mariot17}, \emph{orthogonal arrays}~\cite{mariot18} and \emph{disjunct matrices}~\cite{knezevic18}.

\section{Incremental Construction with EA}
\label{sec:inc-constr}

%We now describe our approach to constructing permutation codes with evolutionary algorithms. 
From an intuitive point of view, it seems natural to cast the search of a permutation code as a combinatorial optimization problem. Given the length $n$ of the permutations, the (minimum) distance $d$ and the number $m$ of desired permutations in the array, one needs to find a set of $m$ elements from $S_n$ such that the Hamming distance between any two permutations in it is at least $d$. Therefore, disregarding the bounds on $m$ induced by the distance parameter, the size of the resulting search space $\mathcal{S}_{m,n}$ is $|\mathcal{S}_{m,n}| = \binom{n!}{m}$ since we need to pick $m$ elements from a set of size $n!$. Exhaustively searching for a solution would be already prohibitive for very small values of $n$ and $m$: for example, there are only $|S_5| = 120$ permutations of length $n=5$. However, visiting all subsets of $S_5$ of size $m=12$ would imply a search space of $\binom{120}{12} \approx 1.05 \cdot 10^{17}$ elements, which clearly cannot be explored in a reasonable amount of time. Consequently, it seems interesting to address this optimization problem with evolutionary algorithms.

\subsection{Evolving Subsets of Permutations}
\label{subsec:subsets}

Given $n, m$ and $d$, a straightforward option is to set up an EA that searches for permutations codes by directly evolving a set of $m$ permutations.
A candidate solution in the population is represented as a matrix $A$ of size $m \times n$, where each row is a permutation of the set $[n]$. Then, this candidate solution would be evaluated through a fitness function that measures how close is $A$ from being a $PA(n,d)$. This could be accomplished, e.g., by counting the number of pairs of rows in $A$ that are at Hamming distance at least $d$ and maximizing such fitness. 
%Concerning how the population is variated and new individuals are introduced in it, one can leverage the extensive literature of genetic operators for permutation-based chromosomes, such as those developed for the \emph{traveling salesperson problem}~\cite{larranaga99}. 
In this approach, one could use common operators for permutation-based chromosomes. For crossover, these include among others \emph{partially mapped crossover}~\cite{goldberg85} and \emph{cycle crossover}~\cite{oliver87}. %, and \emph{edge recombination crossover}~\cite{whitley89}. 
For mutation, the most natural solution is to apply a simple swap operator that randomly exchanges two values in a permutation~\cite{banzhaf90}, but other methods have been proposed such as, e.g., the \emph{inversion operator}~\cite{fogel93} and the \emph{scramble operator}~\cite{syswerda91}. %\\ %More precisely, such operators have been studied for problems where the candidate solutions are encoded by \emph{single} permutation chromosomes.
Still, when evolving permutation codes, one deals with sets of permutations. Thus, a possible solution for this problem would be to apply the variation operators in a \emph{row-wise} manner. For example, given two $m \times n$ arrays $A$ and $B$, define an offspring array $C$ by first applying a permutation-based crossover to the first row of $A$ and $B$, then to the second one, and so on until $C$ is completed. 
%In principle, each row of the offspring array could also be formed by crossing two random rows from the parents instead of proceedings in a uniform way. As we recalled in Section~\ref{sec:bg}, the order of the rows in a $PA(n,d)$ is irrelevant; hence, there is no need to follow the order of the parents' rows. Similarly, one can apply row by row a permutation-based mutation operator to a single array to generate a new mutated individual. This element-wise application of the variation operators is similar to the one adopted by the authors in~\cite{mariot18} to evolve \emph{orthogonal arrays} (OA): in particular, the crossover and mutation operators were applied column-wise in that case.
Although straightforward, this optimization approach suffers from several drawbacks:
\begin{compactitem}
	\item As the aim is constructing a permutation array in an ``all-at-once'' fashion, an EA would directly explore $m$-subsets of the symmetric group $S_n$, which results in a very large search space already for small values of $m$ and $n$.% Consequently, the performances of the EA might not scale well when searching for large permutation arrays.\todo{would be good to give some results after showing it or be more firm in the statement}
	\item The fitness function would need to consider the Hamming distance of each pair of rows in the arrays. Hence, the computational complexity required to evaluate a single candidate solution would be quadratic in the number of permutations of the array, as there are $\binom{m}{2} = \frac{m(m-1)}{2}= \mathcal{O}(m^2)$ pairwise Hamming distances to compute in an $m \times n$ array.
	\item This optimization approach relies on the number of rows $m$ composing the desired permutation to define the problem instance. This implies that one would need to check in advance if $m$ rows are attainable by a permutation code of length $n$ and distance $d$. %, for example, by checking that $m$ lies between two known bounds. Second, 
	%and 2) it would be cumbersome to adopt this optimization approach to address the main question in the field of permutation codes, i.e., determining $M(n,d)$.
\end{compactitem}

\subsection{Iterative Approach}

Given the problems featured by the ``all-at-once'' method, we chose to follow an \emph{iterative} optimization approach, greatly reducing the search space handled at each step by the evolutionary algorithm. Given $n,d \in \N$, the idea is to start from an empty set and add a random permutation $p_1 \in S_n$ of length $n$: trivially, $P$ forms a $PA(n,d)$ with $m=1$ rows. Then, an EA evolves a single permutation $p_2 \in S_n$, until it finds one whose Hamming distance from $p_1$ is at least $d$. When it is found, $p_2$ is added to $P$, thereby expanding the permutation code to $m=2$ rows. The process is repeated by evolving a new permutation until a general termination criterion is met, such as reaching a theoretical upper bound for $M(n,d)$ or a specified number of fitness evaluations. By construction, the obtained array will be a permutation code $PA(n,d)$ with a certain number of rows $m$. At each stage, the EA only explores the set $S_n$ of all permutations of length $n$ instead of the whole set of $m$-subsets of permutations.

More formally, given a $PA(n,d)$ $P = \{p_1, \cdots, p_m\}$ with $m$ rows, the decoding of a candidate chromosome $p_{m+1} \in S_n$ results in the following phenotype: $P' = P \cup \{p_{m+1}\}$. %\todo{I don't think dec() is defined beforehand; or is this a definition?} \todo{I confirm: this is the first time dec() is used, and we have never mentioned phenotypes before}
Clearly, since $P$ already satisfies the properties of a permutation code of minimum distance $d$, the fitness of the candidate solution $P'$ encoded by $p_{m+1}$ is evaluated only by taking into account the $m$ Hamming distances $d_H(p, p_{m+1})$, with $p$ ranging over $P$. This is a much more efficient fitness function since its computational complexity scales linearly with the number of rows in $P$. Further, suppose that $\chi: S_n \times S_n \to S_n$ is a crossover operator for single permutations. Then, given two parent permutations $p_1, p_2 \in S_n$, the phenotype $C$ for the offspring child candidate solution is defined as $C = P \cup \{\chi(c_1, c_2)\}$, that is, crossover is limited only on the new row. Accordingly, the same approach is adopted for mutation by applying the corresponding operator $\mu: S_n \to S_n$ only on the new permutation optimized by the EA.

Algorithm~\ref{alg:ea-inc} reports the pseudocode for the incremental EA informally introduced above. The input parameters are the length of the permutations $n$, the required minimum distance $d$, the fitness budget $fb$, the target number of rows $M$ (which specifies, for example, a known upper bound for $M(n,d)$), the size of the EA population $popsize$, and a vector $\theta$ specifying the parameters for the underlying evolutionary algorithm.

\begin{algorithm}[t]
	\scriptsize
	\floatname{algorithm}{Algorithm}
	\caption{{\sc Incremental-EA-PA}$(n, d, fb, M, popsize, \theta)$}
	\label{alg:ea-inc}
	\begin{algorithmic}
		\STATE $P \gets \{\}$
		\STATE $\pi \gets$ {\sc Gen-Rand-Permutation}$(n)$
		\STATE $P \gets P \cup \{\pi\}$
		\STATE $eval \gets 0$
		\WHILE{$|P| < M$ AND $eval < fb$}
		\STATE $pop \gets$ {\sc Init-Population}$(n, popsize)$
		\STATE {\sc Evaluate-Fitness}$(pop, P, n, d, ev)$
		\STATE $best \gets $ {\sc Update-Best-Ind}$(pop)$
		\WHILE{$eval < fb$ AND (NOT {\sc Is-PA}$(n,d, P, best.\pi)$)}
		\STATE $pop \gets $ {\sc Update-Pop-EA}$(n, d, P, pop,\theta, eval)$
		\STATE $best \gets $ {\sc Update-Best-Ind}$(pop)$
		\ENDWHILE
		\IF{{\sc Is-PA}$(n,d, P, best.\pi)$}
		\STATE $P \gets P \cup \{best.\pi\}$
		\ENDIF
		\ENDWHILE
		\STATE return $P$ 
	\end{algorithmic}
\end{algorithm}

The subroutines {\sc Gen-Rand-Permutation} and {\sc Init-Population} respectively generate at random a single permutation and a population of $popsize$ candidate permutations. An individual $ind$ in the population is assumed to be a record composed of two items, namely $ind.\pi$ (the vector specifying the permutation) and $ind.fit$ (the fitness value of the permutation). 
{\sc Evaluate-Fitness} computes the underlying fitness function for each individual in the population, while {\sc Update-Best-Ind} returns a pointer to the best individual in the current population. The specific structures for these two subroutines depend on the details of the fitness function, which we will address in the next section. {\sc Is-PA} is a predicate returning true if and only if the union of a $PA(n,d)$ and a new permutation is still a $PA(n,d)$, and it is used to determine when to exit from the inner while loop of the EA. When an optimal solution is found, {\sc Is-PA}$(n,d, P, best.\pi)$) returns true, and the permutation code $P$ is extended by adjoining to it the permutation of the best individual in the population.

The actual EA is implemented by the {\sc Update-Pop-EA} subroutine. In particular, depending on the underlying EA, the population might be updated completely, as in a generational approach (possibly coupled with elitism) or only partially, with only a few new offspring individuals entering into the population at each step. In our experiments, we adopted a steady-state genetic algorithm (GA) with tournament selection. This means that each time {\sc Update-Pop-EA} is invoked, $t$ individuals are drawn at random from the population, and the two with the best fitness values are selected for crossover. The resulting offspring then undergoes mutation with probability $p_{\mu}$, that replaces the worst individual in the tournament. Algorithm~\ref{alg:ga-steady} gives the pseudocode for our steady-state GA implementing the {\sc Update-Pop-EA} subroutine. The parameters vector $\theta$ is replaced by the pair $(t, p_{\mu})$, whose components respectively specify the tournament size and the mutation probability. Notice also that $eval$, which is a counter used to keep track of the number of fitness evaluations performed by the algorithm, is assumed to be a global variable: in fact, it is used in the invariants for the while loops in Algorithm~\ref{alg:ea-inc}.

\begin{algorithm}[t]
	\scriptsize
	\floatname{algorithm}{Algorithm}
	\caption{{\sc Update-Pop-EA}$(n, d, P, pop, (t, p_{\mu}), eval)$}
	\label{alg:ga-steady}
	\begin{algorithmic}
		\STATE $tourn \gets$ {\sc Random-Select}$(t, pop)$
		\STATE $(p_1,p_2) \gets ${\sc Select-Best}$(tourn)$
		\STATE $c \gets$ {\sc Crossover}$(p_1,p_2)$ 
		\STATE $c \gets ${\sc Mutation}$(c, p_{\mu})$
		\STATE $c.fit \gets$ {\sc Fitness}$(n,d,P, c)$
		\STATE $eval \gets eval + 1$
		\STATE $worst \gets $ {\sc Select-Worst}$(tourn)$
		\STATE {\sc Replace}$(worst, c)$
		\STATE return $pop$
	\end{algorithmic}
\end{algorithm}
The subroutines {\sc Random-Select}, {\sc Select-Best}, and {\sc Select-Worst} respectively return a random subset of $t$ individuals from the population, the two best individuals and the worst one in the tournament concerning their fitness values. The offspring chromosome is created from $p_1$ and $p_2$ by first applying {\sc Crossover} and then {\sc Mutation}. After evaluating the fitness function -- and increasing the counter of fitness evaluations -- the subroutine {\sc Replace} changes the worst individual in the tournament to the newly created offspring.

\subsection{Fitness Functions}
\label{subsec:fit}

We defined four fitness functions to be optimized by the iterative EA described in the previous section, which we describe below. In what follows, we assume that the goal is to compute the fitness of a permutation $p \in S_n$ when adjoined to a $PA(n,d)$ of $m$ rows, $P = \{p_1,\cdots,p_m\}$.

The first fitness function directly sums the Hamming distances of each pair $(p,p_i)$ of permutations, but only if they are at least equal to the required minimum distance $d$:
\begin{equation}
\label{eq:fit-1}
fit_1(p) = \sum_{p_i \in P} \delta_i \cdot d_H(p,p_i), \textrm{ where } \delta_i = \begin{cases} 1, & \textrm{ if } d_H(p,p_i) \ge d, \\ 0, & \textrm{ otherwise }\end{cases} \enspace .
\end{equation}
Note that this fitness function completely neglects the permutation pairs' information at Hamming distance lower than $d$. For this reason, the second fitness function has the same form of $fit_1$, but also takes into account the invalid pairs by discounting them through an exponential factor:
\begin{equation}
\label{eq:fit-2}
fit_2(p) = \sum_{p_i \in P} \delta_i' \cdot d_H(p,p_i), \textrm{ where } \delta_i' = \begin{cases} 1, & \textrm{ if } d_H(p,p_i) \ge d, \\ 2^{d_H(p,p_i)-d}, & \textrm{ otherwise }\end{cases} \enspace .
\end{equation}
Indeed, when $d_H(p,p_i) < d$ the factor $\delta_i'$ is a number between $0$ and $1$, which decreases as the difference $d_H(p,p_i)-d$ gets smaller. In this way, the more a pair $(p,p_i)$ is closer to the required minimum distance $d$, the more it contributes to the fitness function.

The third fitness function considered in our experiments corresponds to the minimum Hamming distance between $p$ and each permutation in $P$, that is,
\begin{equation}
\label{eq:fit-3}
fit_3(p) = \min_{p_i \in P} \left\{d_H(p,p_i) \right\} \enspace .
\end{equation}
Hence, the permutation $p$ is an optimal solution as soon as $fit_1(p) \ge d$, since this is precisely the characterizing property of a $PA(n,d)$. Although straightforward, this fitness function suffers from a limited range of possible values (especially for small values of $d$), making many candidate solutions very similar. This may hamper, in turn, the EA's ability to exploit specific regions of the search space.

The three fitness functions described up to now are all meant to be maximized as an optimization objective. On the contrary, the fourth fitness function is based on counting the number of pairs that do not meet the minimum distance requirement, clearly with the objective of minimizing them:
\begin{equation}
\label{eq:fit-4}
fit_4(p) = \left| \{ (p,p_i): p_i \in P, \ d_H(p,p_i) < d \} \right| \enspace .
\end{equation}

\section{Experimental Evaluation}
\label{sec:exp-plan}

%\subsection{Problem Instances}

A problem instance for the permutation array problem is defined by the length of the permutation $n$ and the (minimum) distance $d$. To evaluate the suitability of the incremental EA on this problem, one possibility is to compare the maximum number of rows obtained by it for a $PA(n,d)$ and the corresponding lower/upper bounds known in the literature. As far as we are aware, Smith and Montemanni~\cite{smith12} report the most up-to-date table that reports such bounds for $6 \le n \le 18$ and $4 \le d \le 18$. Since $d$ must always be less than or equal to $n$, the total number of problem instances to be tested for a complete comparison is $\sum_{i=3}^{15} i = 117$, which might be unfeasible depending on how much time a single run of the EA takes. Therefore, it makes sense to perform the experiments on a subset of instances, limiting the size of the permutations to $n=10$ and minimum distance $n-2 \le d \le n$. In this way, we get a total of 15 problem instances to test. These instances also have practical relevance in the design of modulation schemes for powerline communications (see, e.g.,~\cite{ferreira00}, where $PA$ of length at most $8$ are considered for this task) and for the design of block ciphers, where $n=8$ is a popular permutation size in the diffusion layers of lightweight block ciphers~\cite{liu16}.
%\todo{weak motivation, are those instances also practically relevant?}
The instances where $n=d$ correspond to the problem of finding a Latin square of order $n$. Furthermore, although the size of the symmetric group $S_n$, for $6\le n \le 10$, is sufficiently limited to be completely explored, recall that the unfeasibility of the exhaustive search approach stems from the fact that we are trying to construct \emph{subsets} of permutations. This already yields a search space of size $\binom{6!}{120} \approx 3.07 \cdot 10^{140}$, for the $PA(6,4)$ instance, and thus it cannot be exhaustively explored. For each considered combination of $n$ and $d$, Table~\ref{tab:sizes} reports the size of the corresponding search space computed as $\binom{n!}{M(n,d)}$ and the corresponding best value known for $M(n,d)$ taken from~\cite{smith12}.
\begin{table}[t]
	\caption{Approximate search space size $\mathcal{S}_{n,d}$ and code size bound $M(n,d)$ for each considered problem instance. Bold values represent non-tight lower bounds.}
	\centering
	\begin{tabular}{cp{1.5cm}p{1.75cm}p{1.75cm}p{1.75cm}p{1.75cm}p{1.75cm}}
		\hline
		$d$\textbackslash $n$ & & 6 & 7 & 8 & 9 & 10 \\
		\hline
		\hline
		\multirow{2}{*}{$n-2$} & $\mathcal{S}_{n,d}$ & $3.07 \cdot 10^{140}$ & $2.31 \cdot 10^{277}$ & $1.81 \cdot 10^{843}$ & $1.20 \cdot 10^{1\,658}$ & $3.83 \cdot 10^{2\,978}$ \\
		& $M(n,d)$ & 120 & {\bfseries 77} & 336 & 504 & 720 \\\hline
		\multirow{2}{*}{$n-1$} & $\mathcal{S}_{n,d}$ & $3.41 \cdot 10^{36}$  & $1.91 \cdot 10^{106}$ & $1.10 \cdot 10^{184}$ & $3.26 \cdot 10^{297}$ & $1.61 \cdot 10^{453}$  \\
		& $M(n,d)$ & 18 & 42 & 56 & 72 & 49 \\\hline
		\multirow{2}{*}{$n$} & $\mathcal{S}_{n,d}$ & $1.89 \cdot 10^{15}$ & $1.63 \cdot 10^{23}$ & $1.73 \cdot 10^{34}$ & $3.01 \cdot 10^{45}$ & $1.10 \cdot 10^{60}$ \\
		& $M(n,d)$ & 6 & 7 & 8 & 9 & 10 \\
		\hline
	\end{tabular}
	\label{tab:sizes}
\end{table}
The search space size decreases as the minimum distance approaches the length, with $n=d$ giving the smallest sizes -- although still not amenable to exhaustive search.

\subsection{Experimental Settings}
\label{subsec:exp-set}

In our experiments, we evaluated our evolutionary approach to construct PA along three different components: namely, the \emph{fitness functions} described in Section~\ref{subsec:fit}, the underlying \emph{search algorithm}, and the adopted \emph{update policy}.
The search algorithm refers to the particular procedure used to select a suitable permutation to be added in the current code, i.e., the content of the while loop at lines 11-12 in Algorithm~\ref{alg:ea-inc}. In this case, we adopted a permutation-based genetic algorithm (which we will refer to as EA in the following) and a simple random search (RS) as a baseline method for comparison. In particular, the RS works by drawing at random a new permutation at each iteration of the while loop, which is subsequently added only if it is at a minimum distance $d$ from all previous permutations. On the other hand, the EA follows the {\sc Update-Pop-EA} steady-state procedure described in Algorithm~\ref{alg:ga-steady}.

The update policy is the strategy by which the algorithm constructs the permutation code. The iterative approach laid out in Section~\ref{sec:inc-constr} is based on the {\sc Incremental-EA-PA} procedure (Algorithm~\ref{alg:ea-inc}), which only expands the code when a new permutation at minimum distance $d$ from all the current ones is found. However, this update policy might easily get stuck in local optima. Intuitively, the size achievable by a permutation array constructed incrementally highly depends on the initial permutations chosen. Therefore, if the search algorithm makes a few ``wrong choices'' initially, it might end up with a relatively small list of permutations that cannot be further expanded. 

For this reason, we also experimented with a \emph{random reset} update policy: if a new permutation satisfying the minimum distance requirement is not found within a given number of fitness evaluations in the inner while loop of Algorithm~\ref{alg:ea-inc}, then some previous permutations -- chosen at random -- are removed from the current code.
Also, the number of permutations to be removed is chosen randomly, but the maximum value is modeled after the cooling policy as employed in simulated annealing~\cite{kirkpatrick83}
Initially, the maximum number of codewords to remove can be as high as one-third of the current $PA$ size ($|P|$) but is then decreased at every subsequent random reset, in order to favor the exploration of the search space at the beginning of the optimization process and its exploitation in the later stages.
The actual number of permutations to be removed is a random value in $\{1, \cdots, r\}$, where $r$ is set as $r = \frac{1}{3} \times |P| \times e^{-evals/10^6}$.
The condition to invoke this reset is defined as the number of successive evaluations without increasing the $PA$ size, e.g., the number of unsuccessful attempts to add a new permutation to the current $PA$.
In our experiments, this number was defined as $\max(n!, 10^5)$; the reasoning behind this is that $n!$ evaluations of the exhaustive search would be enough to find out whether any new permutation can be added to the current $PA$.
%Since this situation can occur during the optimization, but its occurrence cannot efficiently be predicted, 
Consequently, we use this number as the stagnation detection threshold.
Note that the random reset policy can be used with any search algorithm (i.e., any type of population update) and any fitness function.

In what follows, we will denote by EA1 and EA2 the incremental EA equipped respectively with the plain update policy as in Algorithm~\ref{alg:ea-inc} (where new permutations are only added) and with the above random reset update policy. Likewise, RS1 and RS2 will denote the analogous variants of the RS baseline algorithm concerning the update policy.

Prior to the experiments on the selected problem instances, a short tuning phase was performed on problem instance $PA(7,5)$ to estimate the appropriate parameter values for the population size and the mutation rate $p_{\mu}$ of the GA. Based on those results, the population size was set to $1\,000$ individuals, and the mutation rate was kept at $30\%$.
In all the experiments, the total number of evaluations (the fitness budget $fb$) was set to $10^7$, and each experiment was executed in 30 repetitions. Concerning the variation operators, we employed the permutation-based crossovers and mutations implemented in the ECF framework\footnote{Framework available at~\url{http://ecf.zemris.fer.hr/}}, by choosing them uniformly at random at each evaluation.

\subsection{Results}
\label{subsec:results}

\begin{figure}[!ht]
	\centering
	\includegraphics[width=0.5\columnwidth]{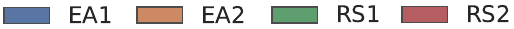}
	\begin{subfigure}{0.5\textwidth}
		\includegraphics[width=\columnwidth]{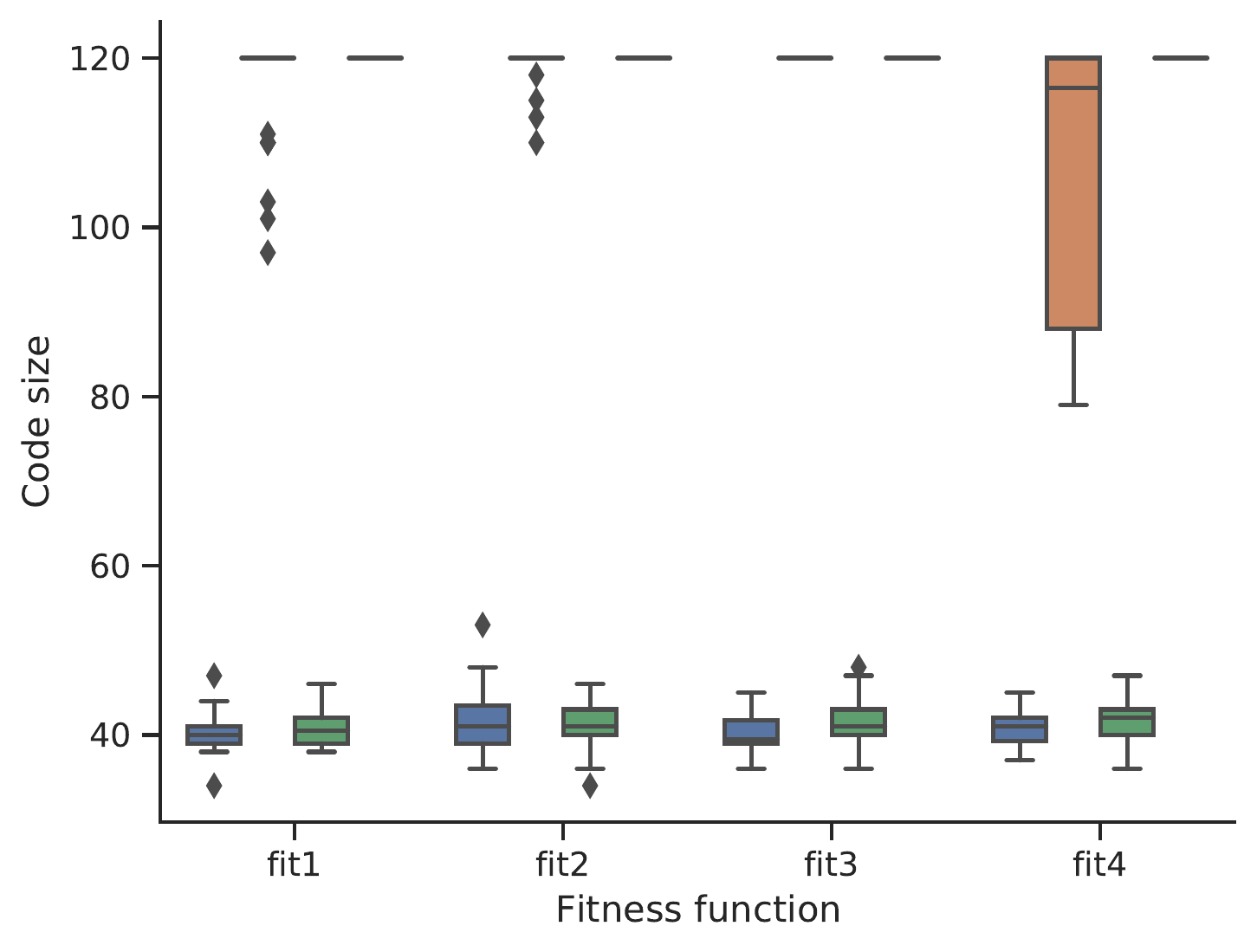}%
		\caption{$6,4$}
	\end{subfigure}%
	\begin{subfigure}{0.5\textwidth}
		\includegraphics[width=\columnwidth]{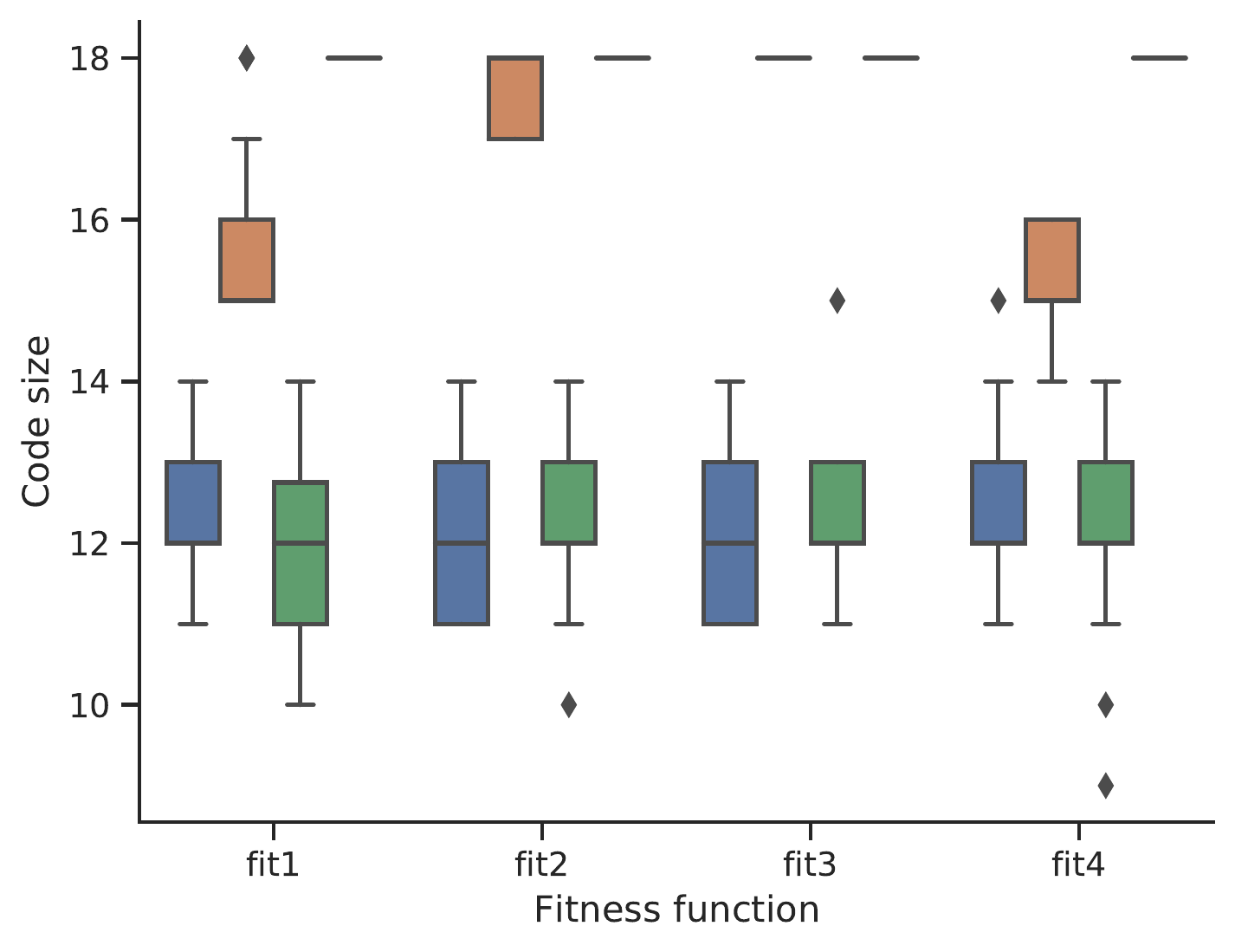}%
		\caption{$6,5$}
	\end{subfigure}
	\begin{subfigure}{0.5\textwidth}
		\includegraphics[width=\columnwidth]{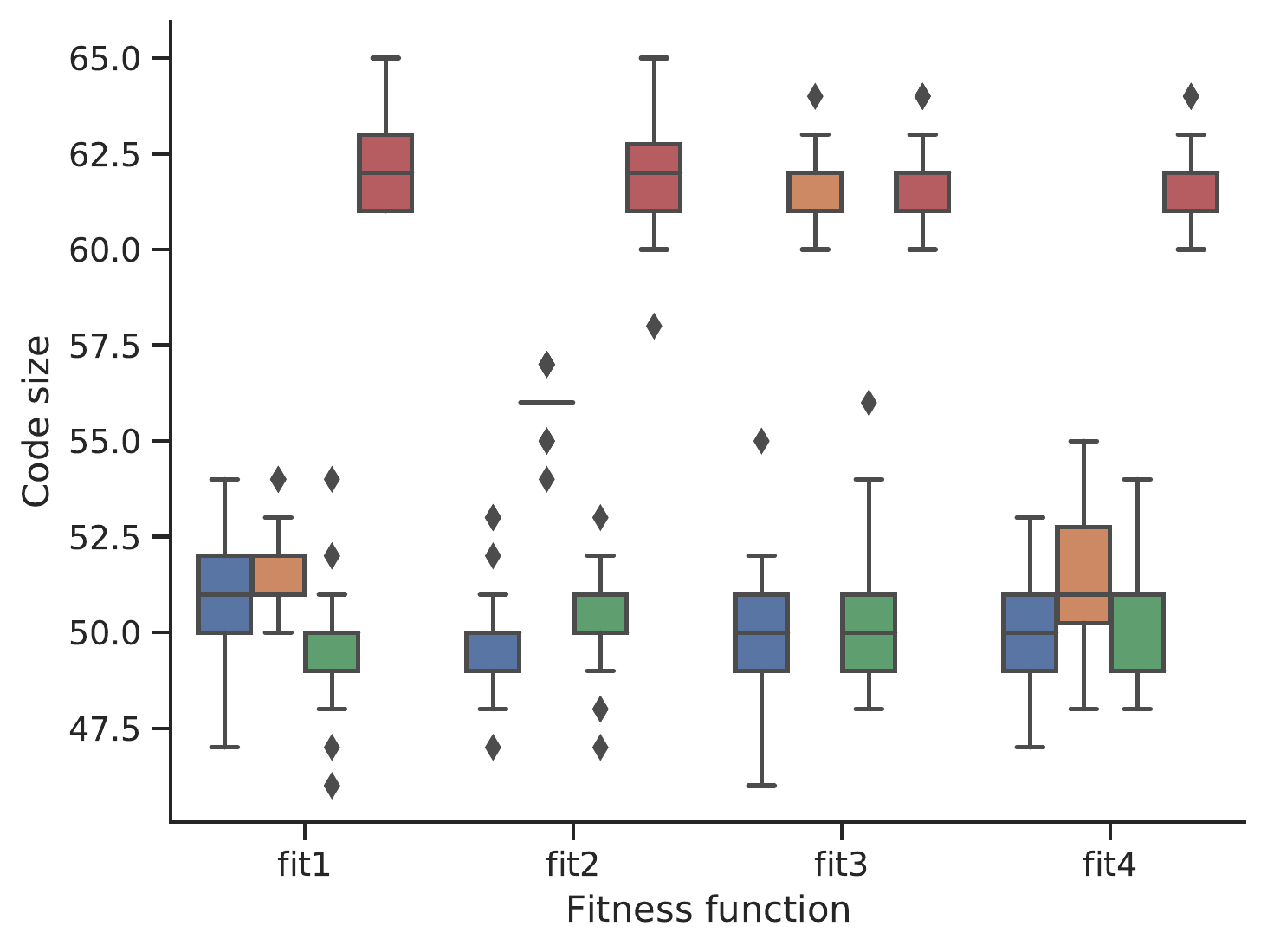}%
		\caption{$7,5$}
	\end{subfigure}%
	\begin{subfigure}{0.5\textwidth}
		\includegraphics[width=\columnwidth]{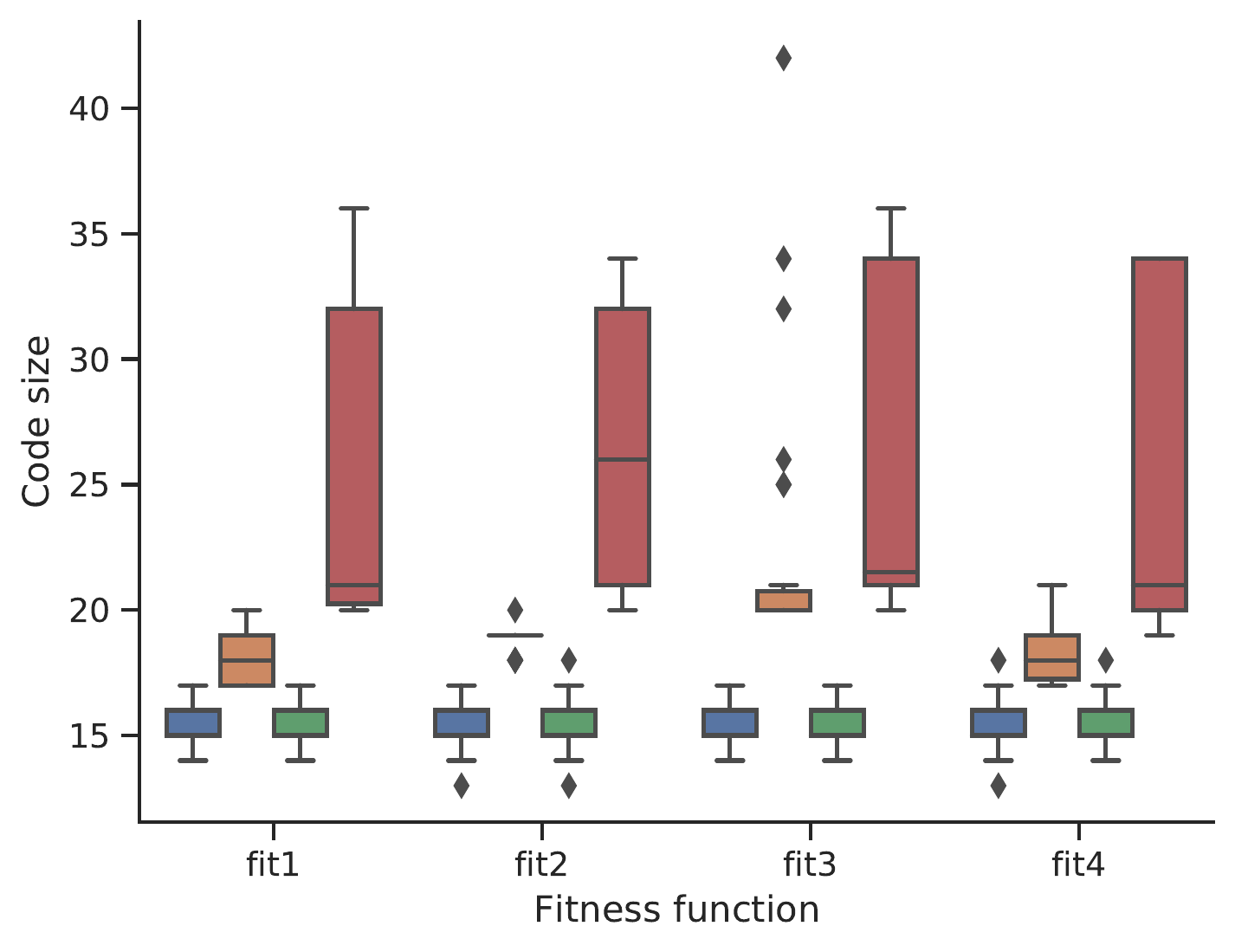}%
		\caption{$7,6$}
	\end{subfigure}
	\begin{subfigure}{0.5\textwidth}
		\includegraphics[width=\columnwidth]{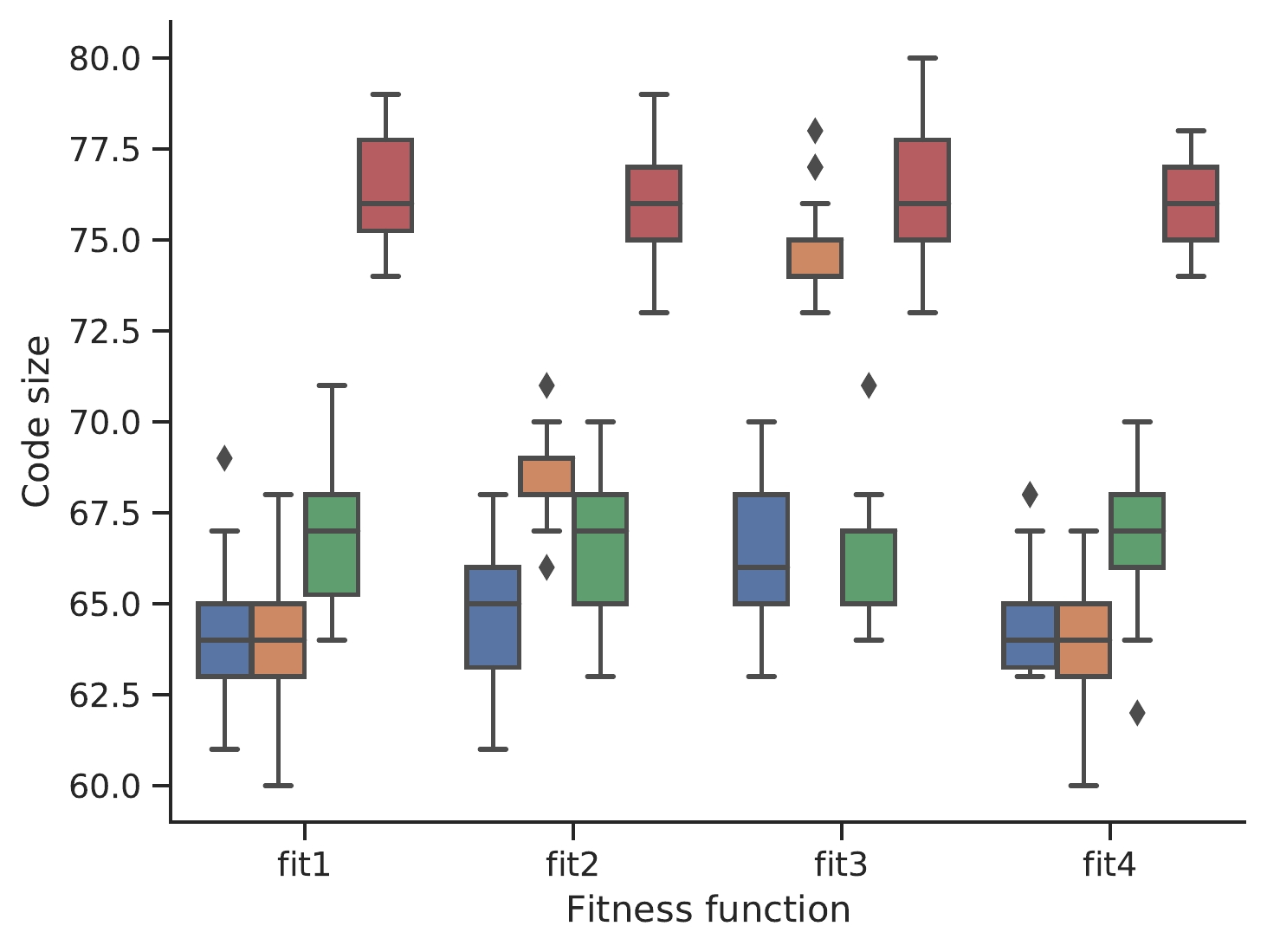}%
		\caption{$8,6$}
	\end{subfigure}%
	\begin{subfigure}{0.5\textwidth}
		\includegraphics[width=\columnwidth]{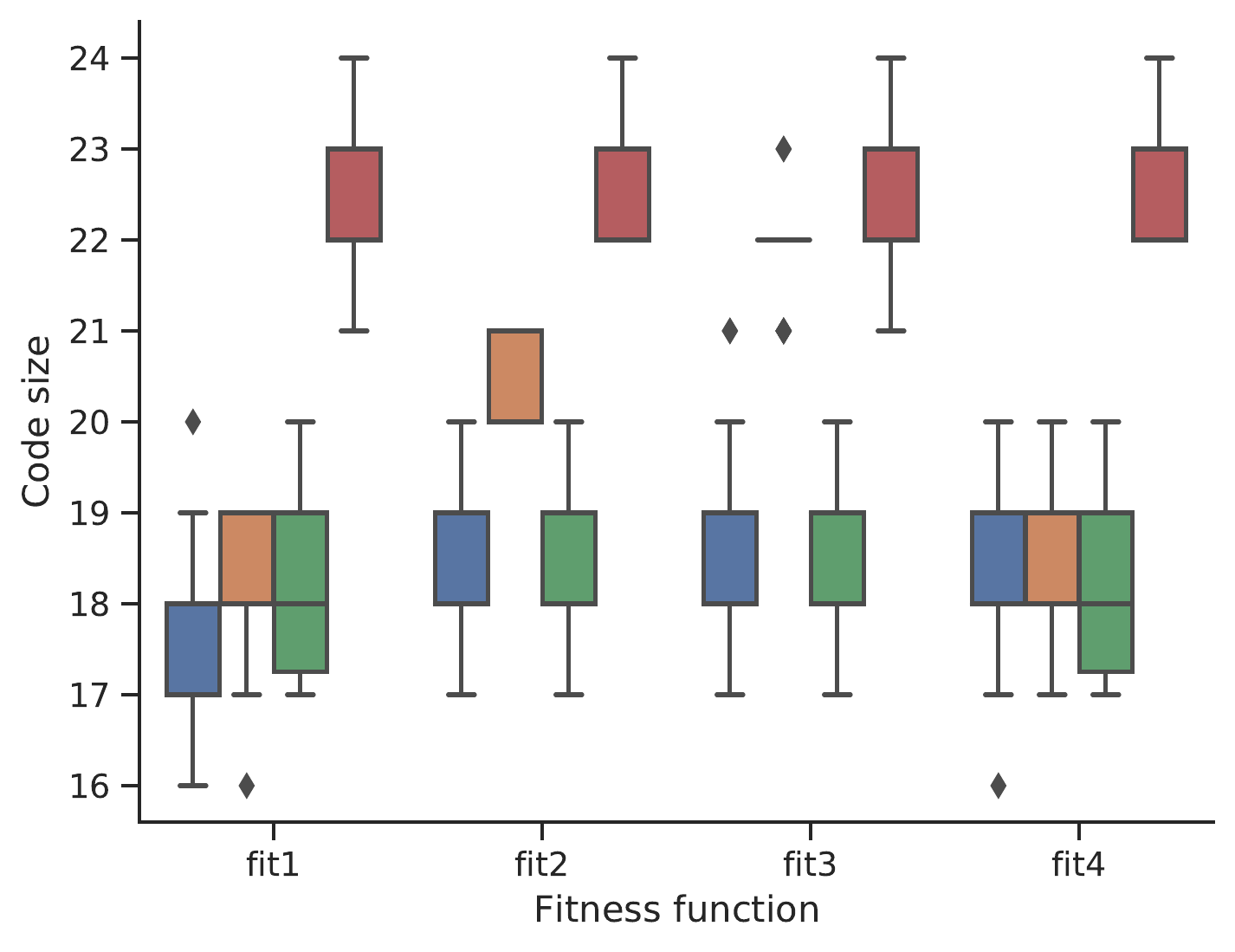}%
		\caption{$8,7$}%
	\end{subfigure}
	\caption{Largest code size achieved by all methods across the problem instances with $n=6,7,8$ and $d=n-2,n-1$.}
	\label{fig:plots-6-8}
\end{figure}

\begin{figure}[!ht]
	\centering
	\includegraphics[width=0.5\columnwidth]{legend.png}
	\begin{subfigure}{0.5\textwidth}
		\includegraphics[width=\columnwidth]{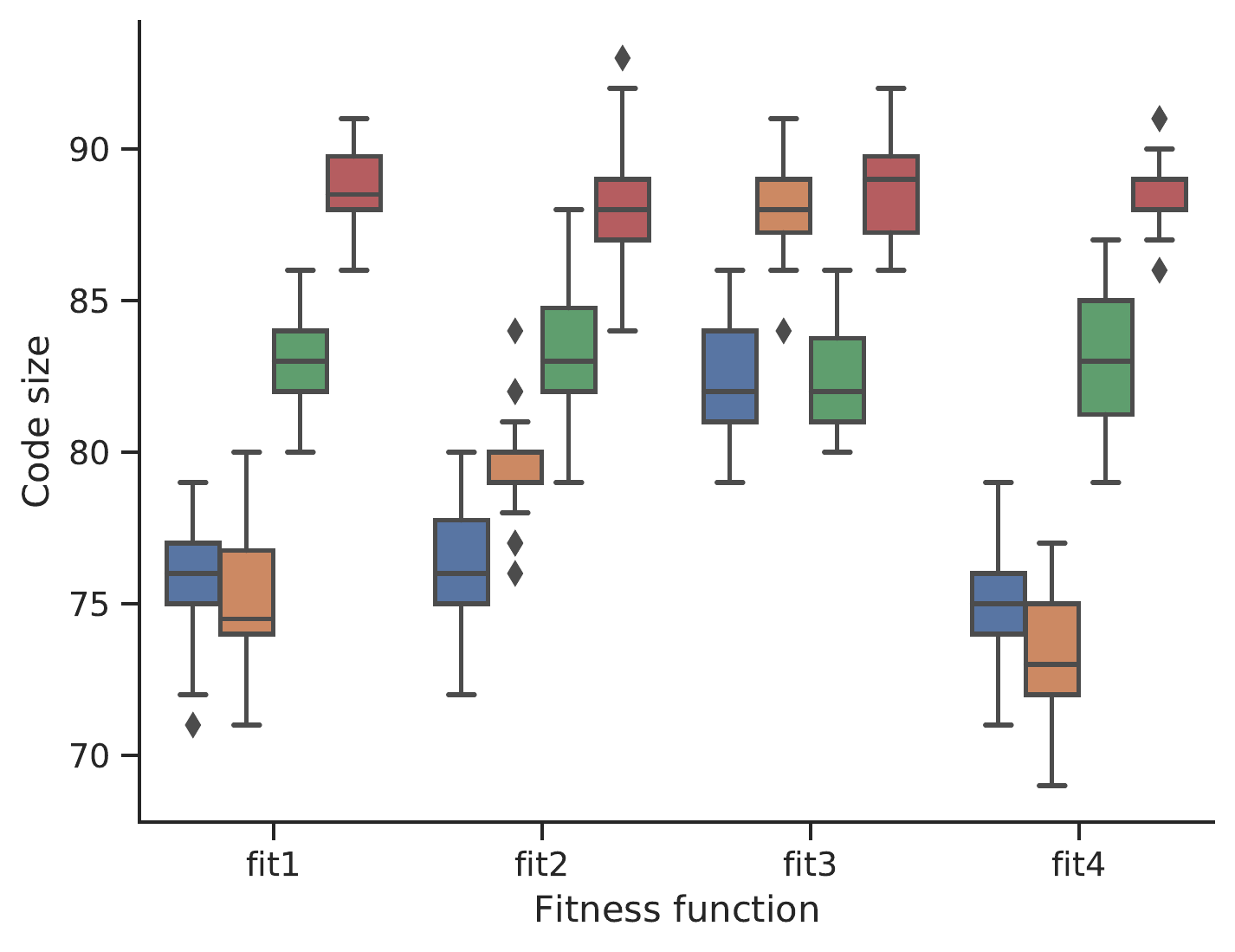}%
		\caption{$9,7$}
	\end{subfigure}%
	\begin{subfigure}{0.5\textwidth}
		\includegraphics[width=\columnwidth]{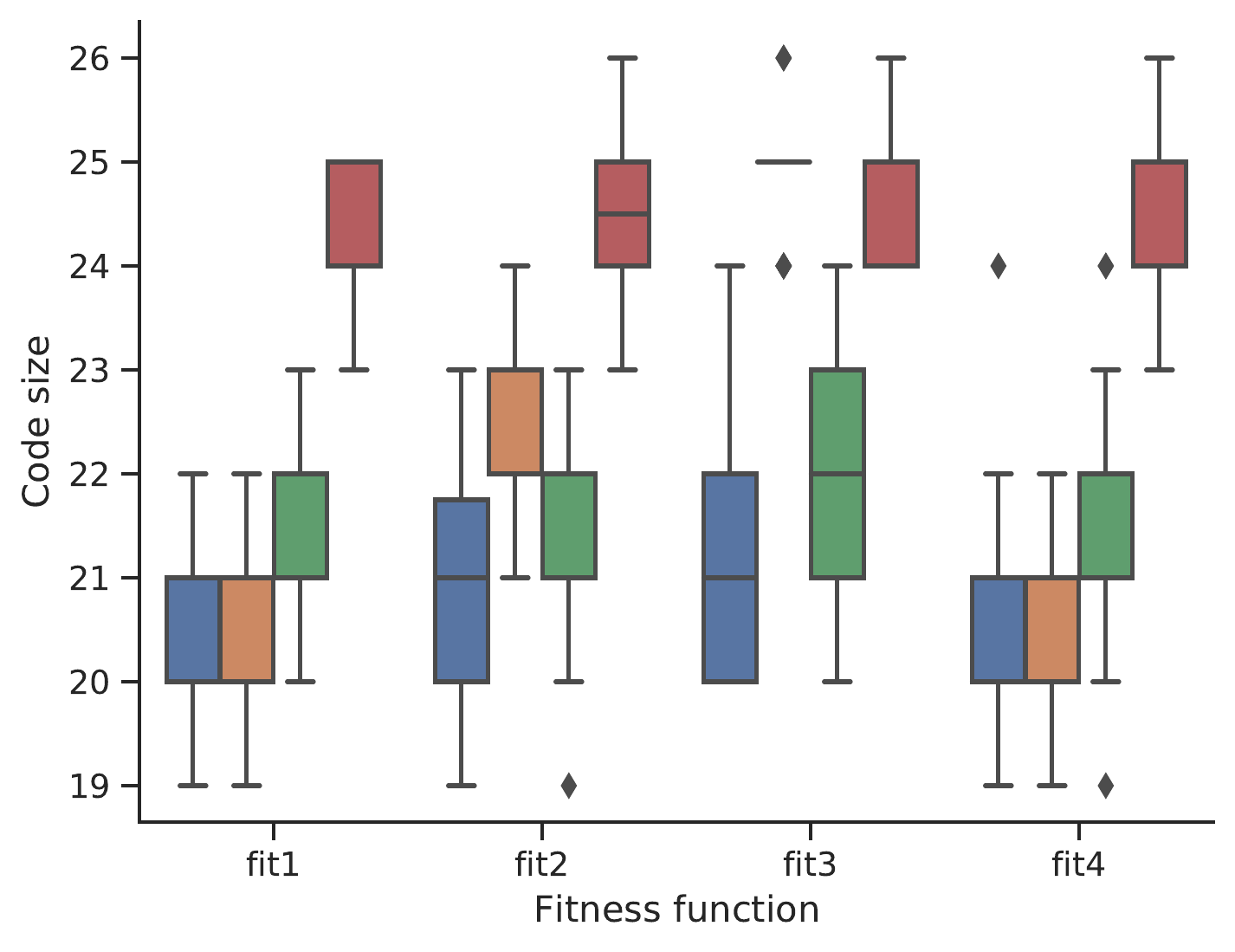}%
		\caption{$9,8$}
	\end{subfigure}
	\begin{subfigure}{0.5\textwidth}
		\includegraphics[width=\columnwidth]{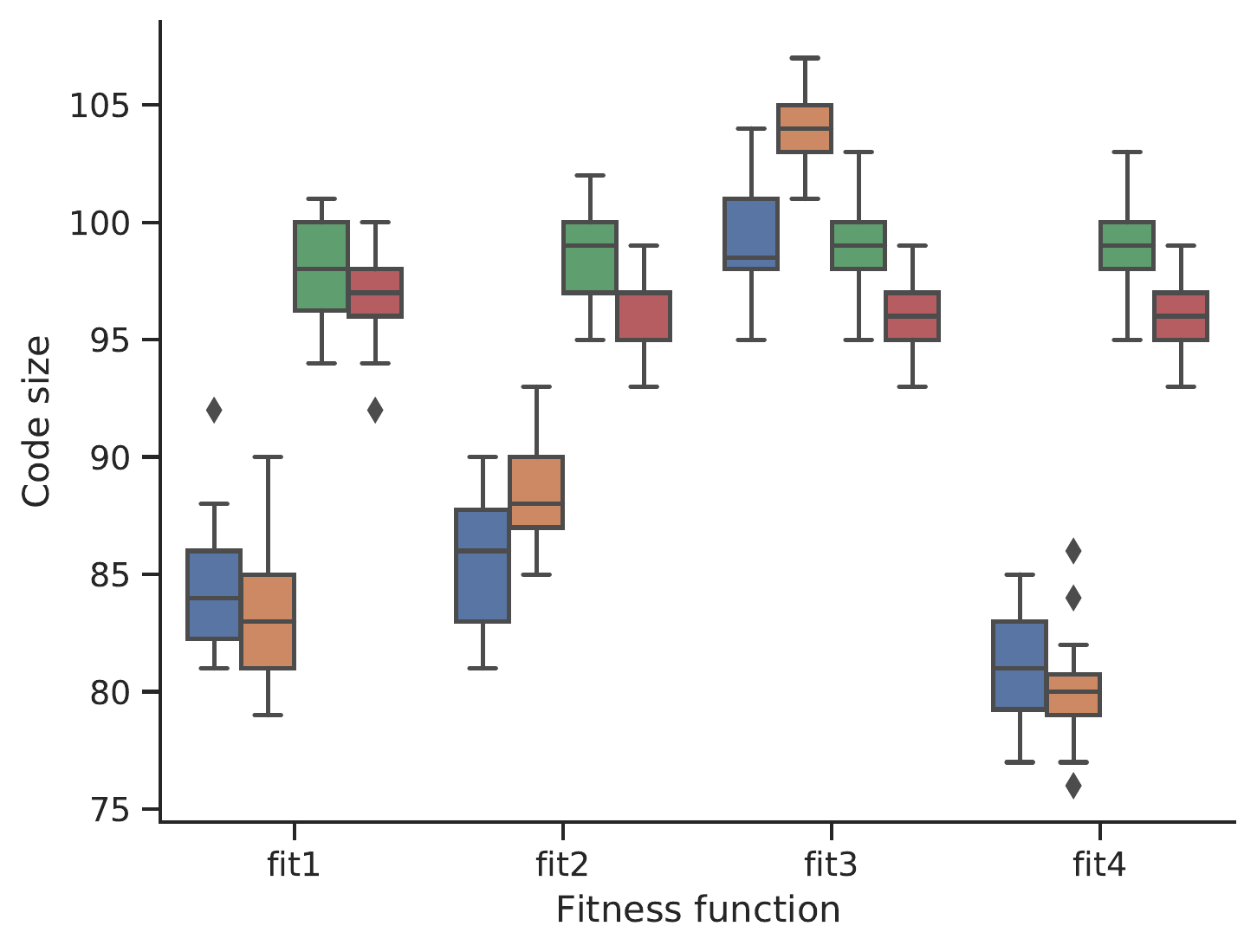}%
		\caption{$10,8$}
	\end{subfigure}%
	\begin{subfigure}{0.5\textwidth}
		\includegraphics[width=\columnwidth]{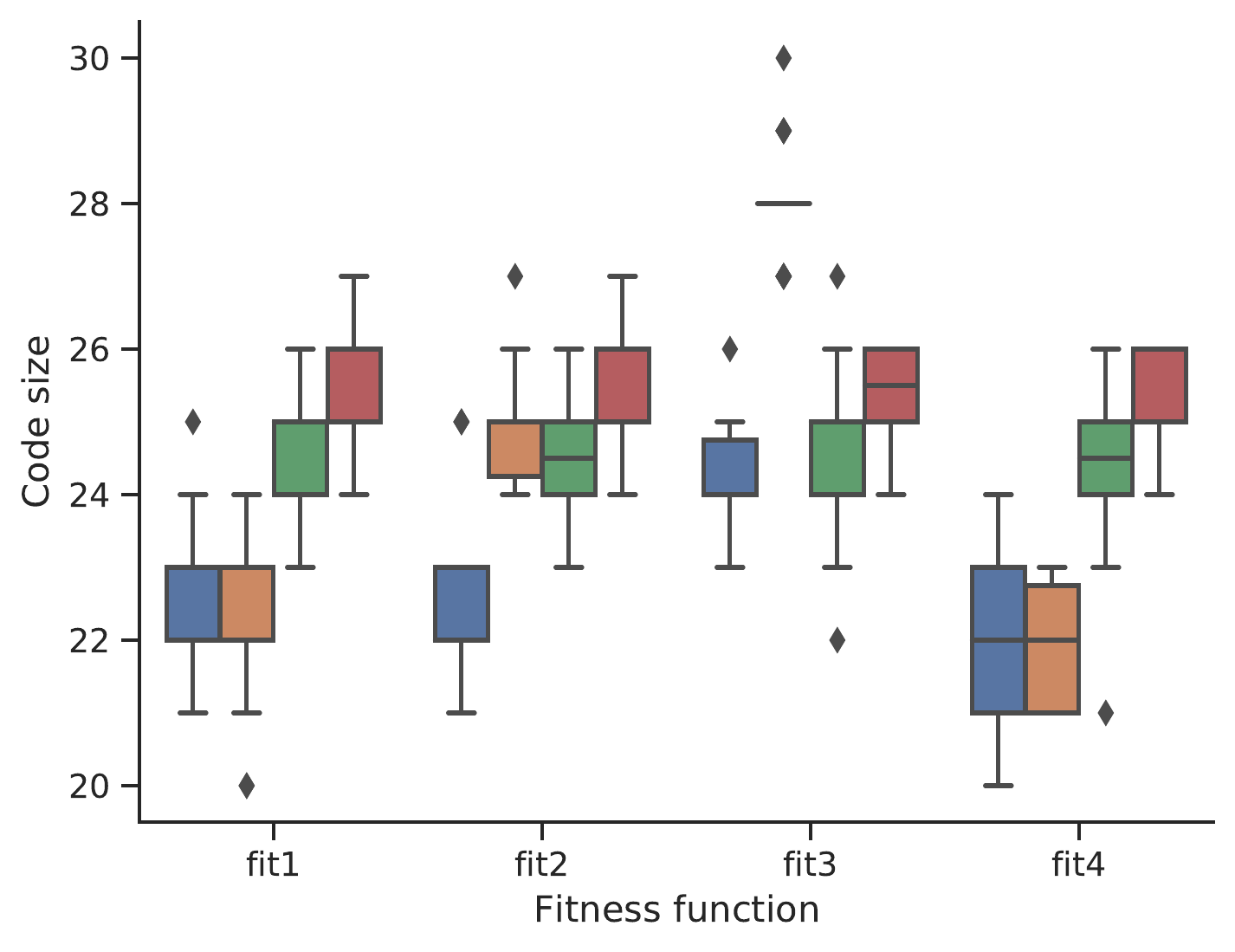}%
		\caption{$10,9$}
	\end{subfigure}
	\caption{Largest code size achieved by all methods across the problem instances with $n = 9, 10$ and $d=n-2,n-1$.}
	\label{fig:plots-9-10}
\end{figure}

Figures~\ref{fig:plots-6-8} and~\ref{fig:plots-9-10} display the results obtained in our experiments with all search methods and across all considered problem instances, except those where $n=d$. In fact, in all those cases, each search variant managed to construct a $PA(n,n)$, or equivalently a Latin square of order $n$. For each problem instance $(n,d)$, the corresponding boxplot shows the four fitness functions against the largest code size achieved by the corresponding variant of a search method. The legend for the four considered combinations of search method and update policy is reported on top of each figure.

First, one can see from the plots that all considered methods, independently from the fitness function, the search method, and the update policy, cannot scale very well concerning the problem size. Indeed, optimal solutions reaching known values for $M(n,d)$ are consistently found only in the $n=6$ case, with the exception of a single $PA(7,6)$ of size $42$ found by EA2 with fitness function $fit_3$. Contrarily, for $n\ge 7$, all considered variants find $PA$ that are significantly smaller than the best-known bounds reported in~\cite{smith12}. Our methods are always able to outperform the Gilbert-Varshamov bound, which is, however, quite loose as reported in Section~\ref{sec:bg}.

As expected, the random reset update policy generally achieves better results than the plain one. This effect is particularly evident in the $n=6$ problem instances, with the combinations adopting the plain update policy obtaining considerably smaller codes than those using random resets, which instead find almost always an optimal solution.
Surprisingly, by comparing the results concerning the update policies, there is no significant difference between the code sizes obtained by EA and RS. A second surprising remark concerns the fitness functions: while we expected $fit_3$ to be the worst-performing one in Section~\ref{subsec:fit}, it generally achieved larger code sizes than the other three.

% \subsection{Discussion}
% \label{subsec:disc}

%We now discuss the results obtained in our experiments, attempting an interpretation of the empirical findings pointed out in the previous section.
%Recall that our experimental evaluation unfolds along three main factors: the four fitness functions, the two search methods (EA and RS), and the two update policies applied to them (plain and random reset).\todo{do we need these two sentences? they say nothing new}

As for fitness functions, the most interesting remark is that the best performing one is also the simplest, namely $fit_3$, that measures the minimum distance of the new candidate solution from all permutations in the current code. This is a fairly straightforward translation of the property characterizing a permutation code into an objective function to be maximized, and we hypothesized that it could underperform due to its limited range of values. The reason why $fit_3$ achieves the largest code sizes over all considered instances might reside in the size of the ``local'' search space, i.e., in the set of all permutations of size $n$ that the EA searches at each stage of the incremental construction. Indeed, we targeted relatively small permutations, where the symmetric group is composed of at most $10! \approx 3^\cdot 10^6$ in the largest considered instance. Using finer-grained fitness functions such as $fit_1$, $fit_2$, and $fit_4$ might have hampered the EA search process, investing many fitness evaluations in optimizing much more information -- e.g., the discounted invalid distances in Eq.~\eqref{eq:fit-2} -- than what was needed. It could be interesting to perform experiments on larger instances where the symmetric group $S_n$ is not amenable to exhaustive search to see if this trend continues or if the additional information exploited by the other fitness functions gives an advantage over $fit_3$.

The relatively small size of the local search space of permutations might also be related to the substantial equivalence of the EA and RS performances. It could be the case that the results are quite similar because it does not make any difference how the local permutation is selected to incrementally expand the code, given the size of the underlying symmetric group. Thus, in this setting, the update policy seems to be the factor that mainly influences the size of the largest codes found by our incremental approach, independently of the search method. Again, this explanation should be tested against further experiments on larger problem instances. One can see already from the plots in Figure~\ref{fig:plots-9-10} that a difference does arise for $n=10$, with EA2 under fitness function $fit_3$ achieving larger codes than those obtained by RS2. It is reasonable to assume that with larger permutation sizes, the evolutionary approach takes over the random search, with an increasing gap between the two methods.

Besides the comparison with random search, the most interesting observation arising from our experiments is that this optimization problem seems to be exceptionally difficult for evolutionary algorithms. Only for the smallest instances of $PA(6,4)$ and $PA(6,5)$ could we obtain optimal solutions concerning the code size (neglecting the outlier found for $PA(7,6)$). In all other cases, the largest code found (either with EA or RS) always lies far from the best lower bounds for $M(n,d)$. This finding could be interpreted in view of the {\sc Max-Clique} formulation of the problem~\cite{montemanni16}. Suppose we have a graph where the nodes are the permutations in $S_n$, and two nodes are connected by an edge if and only if their Hamming distance is at least $d$. Then, constructing the largest permutation code $PA(n,d)$ is equivalent to searching the largest clique in such a graph. With the incremental construction, one starts from a single node in the graph and then tries to expand as much as possible the clique(s) to which this node belongs. Our evolutionary algorithm, on the other hand, does not take into account the \emph{topology} of this graph, which involves both the region where the initial permutation is located and its \emph{neighborhood}, i.e., the set of its adjacent nodes. This problem might also be worsened because EAs are population-based methods. Hence, by starting with a set of candidate solutions generated at random, one might waste many fitness evaluations to ``move'' the population close to the neighborhood of the clique constructed up to that point. One strategy to cope with this issue could be to experiment with smaller population sizes or to integrate the EA with a local search step that also considers the graph representation.

\section{Conclusions and Future Work}
\label{sec:conclusions}

This paper addresses the optimization problem of constructing permutation codes using EA, which, as far as we know, has not been addressed before. The main question in this domain concerns finding the largest code size for a given permutation length $n$ and a minimum distance $d$. We have developed an incremental construction approach, starting from a single random permutation chosen at random and then using an EA to iteratively expand the code iteratively. We evaluated our method with four fitness functions using two different update policies and in comparison to a baseline random search algorithm. Most importantly, the results of our experiments show that this optimization problem is particularly difficult for evolutionary techniques, with the largest codes found by our EA lying far from the best-known lower bounds in most of the considered problem instances. Further interesting findings include the fact that the simplest fitness function performed the best and that the update policy seems to be crucial for finding large codes rather than the underlying search method.

In future work, we plan to improve our incremental EA approach by following the directions outlined above: experimenting with larger problem instances and including a local search optimization step. We also envision investigating a concept closely related to equidistant permutation codes, where the Hamming distance between codewords must be exactly equal to $d$. Equidistant permutation codes are thus a subset of permutation codes, making our iterative procedure applicable. However, since equidistant permutation codes are more rare, we believe this problem to be even harder.

\bibliographystyle{abbrv}
\bibliography{bibliography}

\end{document}